\documentclass[10pt,twocolumn,letterpaper]{article}
 
\usepackage{cvpr}              
\usepackage{float}
\definecolor{cvprblue}{rgb}{0.21,0.49,0.74}
\usepackage[pagebackref,breaklinks,colorlinks,allcolors=cvprblue]{hyperref}
\usepackage{multirow}
\usepackage[table]{xcolor}
\usepackage{array}
\usepackage{booktabs}
\usepackage{tabularx}
\usepackage{graphicx}   
\def\paperID{*****} 
\def\confName{CVPR}
\def\confYear{2026}

\title{Accelerating Diffusion Decoders via Multi-Scale Sampling and One-Step Distillation}

\author{
Chuhan Wang \\
University of California San Diego \\
\texttt{chw136@ucsd.edu}
\and
Hao Chen\\
Carnegie Mellon University\\
\texttt{haoc3@andrew.cmu.edu}
}

\begin{document}
\maketitle
\footnotetext{Preprint.}
\vspace{2pt}

\footnotetext{Code is available at \url{https://github.com/wangchuhan703/multistage_tokenizer_distillation}.
}

\begin{figure*}[h]
\centering
\includegraphics[width=2.1\columnwidth]{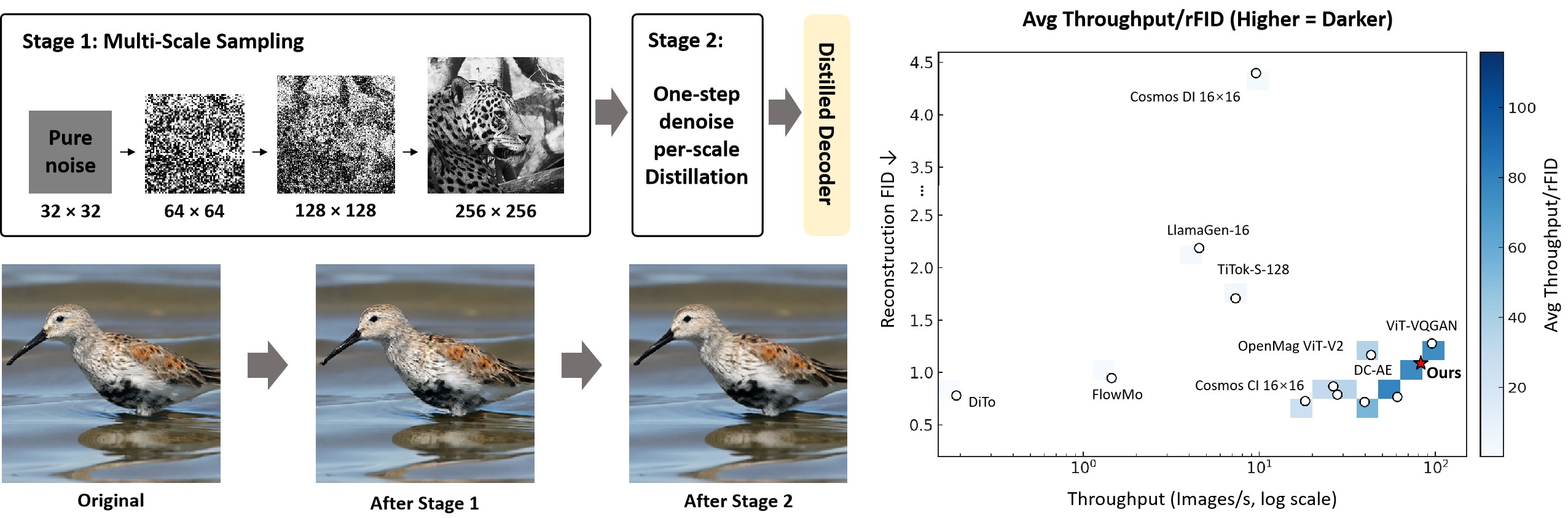} 
\caption{Left: Our two-stage framework reconstructs images through coarse-to-fine sampling and single-step denoising at each scale. Right: Comparison of image tokenizers on rFID and log throughput; shading indicates the throughput-to-rFID ratio. Our method (red star) delivers state-of-the-art efficiency while maintaining strong reconstruction fidelity.}
\label{fig:teaser_fig}
\end{figure*}

\begin{abstract}
Image tokenization plays a central role in modern generative modeling by mapping visual inputs into compact representations that serve as an intermediate signal between pixels and generative models. Diffusion-based decoders have recently been adopted in image tokenization to reconstruct images from latent representations with high perceptual fidelity. In contrast to diffusion models used for downstream generation, these decoders are dedicated to faithful reconstruction rather than content generation. However, their iterative sampling process introduces significant latency, making them impractical for real-time or large-scale applications. 
In this work, we introduce a two-stage acceleration framework to address this inefficiency. First, we propose a multi-scale sampling strategy, where decoding begins at a coarse resolution and progressively refines the output by doubling the resolution at each stage, achieving a theoretical speedup of $\mathcal{O}(\log n)$ compared to standard full-resolution sampling. Second, we distill the diffusion decoder at each scale into a single-step denoising model, enabling fast and high-quality reconstructions in a single forward pass per scale. 
Together, these techniques yield an order-of-magnitude reduction in decoding time with little degradation in output quality. Our approach provides a practical pathway toward efficient yet expressive image tokenizers. We hope it serves as a foundation for future work in efficient visual tokenization and downstream generation.
\end{abstract}

\section{Introduction}
Diffusion autoencoders~\cite{preechakul2022diffusion, chen2025diffusion} have recently emerged as a compelling alternative to traditional autoencoders as image tokenizers for downstream generative models. 
Unlike standard supervised autoencoders that rely on combinations of heuristic losses such as L1 ~\cite{chai2014root},  Learned Perceptual Image Patch Similarity (LPIPS)~\cite{zhang2018unreasonable}, or adversarial loss ~\cite{goodfellow2014generative}, diffusion autoencoders adopt a probabilistic formulation where the decoder reconstructs images by gradually denoising from noise to data ~\cite{ho2020denoising}. 
This formulation enables more expressive and perceptually accurate reconstructions, especially for structured content like text, fine edges, and high-frequency textures ~\cite{dhariwal2021diffusion, rombach2022high, sauer2024fast}. Furthermore, diffusion decoders can naturally model multi-modal distributions, offering robustness in challenging or ambiguous visual scenarios. These advantages have made diffusion autoencoders an increasingly popular choice for modern image tokenizers used in two-stage generation pipelines.

However, these benefits come at a considerable cost: diffusion-based decoders require tens to hundreds of iterative denoising steps to generate a single reconstruction from the latent space, resulting in prohibitively slow inference speeds. This inefficiency limits the practicality of diffusion-based tokenizers in real-time or resource-constrained applications and raises an important question: can we retain the perceptual fidelity of diffusion decoding while substantially improving its efficiency?

\begin{figure*}[t]
    \centering
    \includegraphics[page=1, clip, trim=0 0 0 0, width=\linewidth]{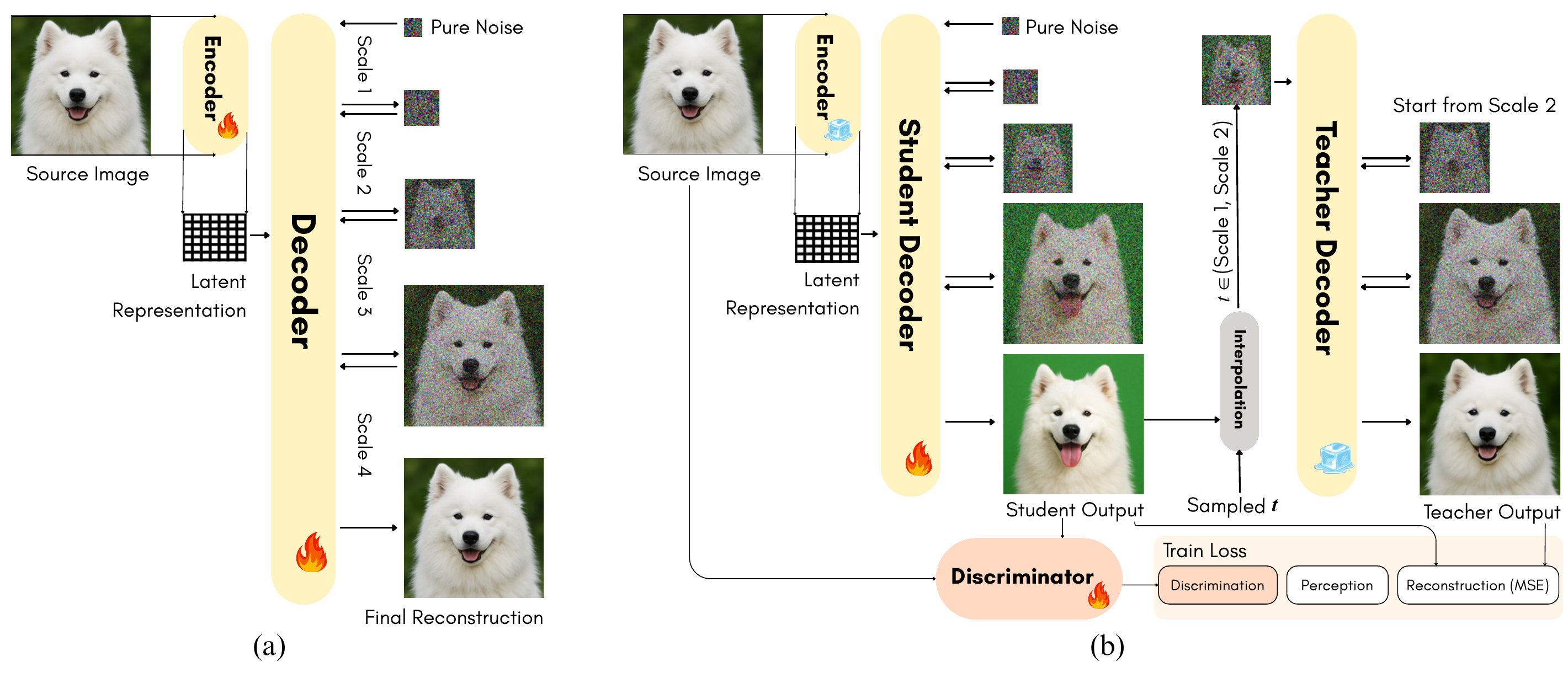}
    \vspace{-2mm}
    \caption{Overview of our two-stage acceleration framework for diffusion decoding. (a) In Stage 1, the decoder progressively reconstructs the image through multi-scale denoising, starting from pure noise at low resolution and upsampling through four spatial scales to obtain a final reconstruction. (b) In Stage 2, this trained decoder is used as the teacher model to supervise a student decoder that performs single-step denoising at each scale. The student is trained with guidance from the teacher outputs, an auxiliary discriminator, and perceptual and reconstruction losses, all conditioned on the same latent representation encoded from the input image.}
    \label{fig:overview}
    \vspace{-2mm}
\end{figure*}

In this work, we propose a two-stage framework that substantially accelerates diffusion decoders in image tokenization pipelines, while preserving their high reconstruction quality. Our first contribution is a \textit{multi-scale sampling strategy}: instead of decoding the image entirely at full resolution, we begin generation at a low-resolution image and progressively double the resolution across a logarithmic number of stages (see Figure~\ref{fig:overview}). This coarse-to-fine decoding scheme leverages the insight that global structure can be synthesized early, while finer details can be refined later. As a result, our method reduces overall computation and achieves decoding complexity of $O(\log n)$ with respect to resolution. We illustrate the inference speed advantage of our method over other tokenizers in Figure~\ref{fig:teaser_fig}.

Even with reduced computation, diffusion-based decoder still requires multiple inference steps to denoise an image.
To further accelerate inference, we introduce a model distillation approach based on a single-step adversarially guided denoising strategy~\cite{sauer2024adversarial}. For each resolution scale, we distill a compact single-step denoiser to approximate the effect of the full multi-step diffusion process over that scale. During inference, these distilled student models replace the original iterative diffusion decoder, allowing the entire image to be reconstructed using only a few forward passes—one per scale. Our distillation framework is inspired by prior adversarial distillation methods~\cite{sauer2024fast, sauer2024adversarial}, where a pre-trained teacher provides denoising supervision and a discriminator ensures perceptual fidelity. In contrast to existing approaches that operate only at a fixed full resolution, our method integrates this distillation into a multi-scale decoding framework, enabling efficient coarse-to-fine reconstruction while remaining compatible with standard diffusion training procedures.

We evaluate our accelerated decoder on multiple image reconstruction benchmarks, and show that it achieves competitive visual fidelity with a significant reduction in decoding time. Our contributions are as follows:
\begin{itemize}
\item We introduce a multi-scale sampling framework for diffusion tokenizers, achieving logarithmic decoding complexity in resolution. In our implementation, decoding proceeds from $32\times32$ to $256\times256$ in 4 stages, yielding up to \textbf{10$\times$} speedup over full-resolution sampling.

\item We propose a novel per-scale distillation method that approximates multi-step diffusion with single-step denoisers at each inference resolution, effectively reducing sampling steps from 50–100 to only 4 in total and further speeding up the reconstruction by \textbf{31$\times$}.

\item We demonstrate that our framework achieves competitive reconstruction quality with significantly improved inference speed across standard visual quality metrics.
\end{itemize}

\section{Related Works}

\subsection{Image Tokenization}
Image tokenization refers to the process of mapping high-dimensional visual data into compact latent representations, letting downstream generative models run with reduced computational complexity. These tokenizers enable high-quality image reconstruction and synthesis while alleviating the burdens of pixel‑space modeling.

Early approaches include continuous tokenizers such as standard Variational Autoencoders (VAEs)~\cite{kingma2013auto} and their variants~\cite{sohn2015learning, higgins2017beta, gregor2018temporal}, which encode images into smooth continuous embeddings optimized via reconstruction loss. Later, discrete codebook-based models such as VQ-VAE~\cite{van2017neural} and VQGAN~\cite{esser2021taming} introduced quantized latent spaces that enabled more effective discrete autoregressive modeling, at the cost of limited expressiveness and occasional codebook collapse. More recently, transformer-based tokenizers like TokenCritic~\cite{lezama2022improved} and TiTok~\cite{yu2024image} further propose to represent images using extremely compact token sequences and rely on autoregressive or masked prediction objectives to learn semantic token representations.
\subsection{Diffusion Autoencoders}
Diffusion autoencoders~\cite{preechakul2022diffusion, chen2025diffusion} combine a deterministic encoder that produces a semantic latent representation with a stochastic diffusion decoder that adds detail. They deliver excellent reconstruction quality and support interpolation and editing. Although latent diffusion decoders achieve high-fidelity reconstruction with fewer steps than conventional diffusion models, their iterative sampling still limits efficiency in real-time or large-scale applications.

Recent multi-scale methods~\cite{chen2025pixelflow, xu2025msf, li2025predicting, skorokhodov2025improving, zhang2024multi} offer a promising direction to improve efficiency. They allocate computation across resolutions, performing coarse denoising at low scales and refining details only at higher resolutions. These approaches significantly reduce inference cost while maintaining generation quality.

We extend the multi-scale idea to diffusion tokenizers by training a pixel decoder that refines images in a coarse-to-fine manner with a single denoising step at each scale. This hierarchical process must co-adapt with the encoder, so we train the decoder entirely from scratch. Pre-trained full-resolution decoders~\cite{zhao2024epsilon, wang2025selftok} are incompatible with the progressive design. 
The resulting tokenizer sharply reduces latency while preserving latent semantics and visual fidelity.




\subsection{Diffusion Model Distillation}
Despite exceptional image quality, diffusion models are inherently slow due to their iterative sampling process, often taking tens to hundreds of steps. To speed them up, many recent efforts have focused on distillation into fast student models~\cite{salimans2022progressive, meng2023distillation, luo2023latent, sauer2024adversarial, zhou2025few}, enabling generation in under 8 steps but often compromising on detail or realism. A leading approach among these is Adversarial Diffusion Distillation (ADD)~\cite{sauer2024adversarial,poole2022dreamfusion} with adversarial training. By combining a distillation loss derived from a frozen teacher diffusion model and an additional adversarial discriminator loss, ADD enables high-fidelity image generation in just 1–4 sampling steps. It outperforms other few-step methods in fidelity and even matches or surpasses state-of-the-art teachers, like SDXL~\cite{podell2023sdxl},  at 4 steps, making real-time, high-quality synthesis practical.

Building on this, we extend the ADD paradigm to multi-scale diffusion autoencoders, where the encoder is frozen and only the decoder undergoes distillation. Our goal is to accelerate denoising process, conditioned on latent inputs from the encoder, reducing decoding steps to just one step per scale. The distilled decoder preserves reconstruction quality, latent semantics, and downstream generation capabilities, while delivering over 30× reduction in decoding latency, closely matching teacher-model performance. 

\section{Preliminaries}
\paragraph{Traditional Tokenizers.}  
Let $E$ and $D$ be an encoder–decoder pair that maps an image
$I\!\in\!\mathbb{R}^{H\times W\times 3}$ to a latent tensor
$Z = E(I)\in\mathbb{R}^{h\times w\times d}$ with
$h = H/f,\; w = W/f$ under a spatial down-sampling factor $f$. \textit{Continuous tokenizers} optimize an evidence lower bound (ELBO) consisting of a pixel reconstruction loss and a KL regularization term on a Gaussian latent space. Variational Auto-Encoders (VAEs) pioneered this approach, and Masked Auto-Encoders (MAEs) later showed that an asymmetric ViT encoder paired with a lightweight decoder can recover masked pixels using only an $L_{2}$ loss\cite{he2022masked,van2017neural}. \textit{Discrete codebook tokenizers} replace the Gaussian prior with a learned dictionary and employ additional commitment and perceptual losses, sometimes combined with an adversarial objective, to better preserve high-frequency details at high compression ratios \cite{van2017neural,esser2021taming}.

To further reduce token budgets, \emph{1-D tokenizers} abandon the
2-D grid entirely: TiTok flattens an image into a sequence of just 32 tokens
and cuts decoding latency by two orders of magnitude without hurting
FID, while TokenCritic adds an auxiliary network that flags low-quality
tokens for resampling in non-autoregressive generation
\cite{yu2024image,lezama2022improved}.
We group these variants under “traditional tokenizers" because their
decoders remain fully deterministic and are trained with pixel or GAN-style
objectives rather than stochastic denoising dynamics.

\paragraph{Diffusion Decoders.}
Diffusion models corrupt data via a forward noising process
\begin{equation}
  q\bigl(x_t \mid x_{0}\bigr)=
  \mathcal{N}\!\bigl(\alpha_t x_{0}, \sigma_t^{2}I\bigr),
  \quad t\in[0,1],
\end{equation}
and optimise a denoising score-matching objective
\begin{equation}
  \mathcal{L}_{\text{DM}}(x_{0}) =
  \mathbb{E}_{t,\varepsilon\sim\mathcal{N}(0,I)}
  \bigl[\lVert \varepsilon_{\theta}(x_t,t) - \varepsilon \rVert_{2}^{2}\bigr],
  \label{eq:diffusion-loss}
\end{equation}
which can be interpreted as an ELBO weighted over log-SNR levels
\cite{ho2020denoising,kingma2023understanding}.
Flow Matching generalises this view by regressing continuous
probability-path vector fields and enables faster ODE sampling paths
\cite{lipman2022flow}.

Replacing $x_{0}$ with the encoder’s latent $Z$ turns the same
objective into a \emph{diffusion tokenizer}.
DiTo shows that a \emph{single} diffusion $L_{2}$ loss suffices to learn
ultra-compact tokens without using GAN or perceptual terms, while remaining
compatible with modern diffusion samplers and supporting one-step
per-scale distillation for real-time decoding \cite{chen2025diffusion}.
These properties motivate our choice of diffusion decoders, elaborated
in the following section.

\section{Method}


Traditional tokenizers compress an image into only a handful of latent tokens and then rely on a single deterministic decoder pass to rebuild the full-resolution picture. With so little information, the decoder must guess global structure and fine texture at once, often producing blurred details or blocky artefacts. Researchers try to fix this by adding GAN discriminators and perceptual losses, but those extra terms are notoriously sensitive: a slight weight change can swing outputs from “over-smoothed” to “ringing with noise.” Because this one-shot decoder cannot refine its guess over multiple steps, any remaining errors stay locked in.

To overcome these limits, we adopt a diffusion decoder. Its multi-step, probabilistic denoising process refines coarse predictions gradually, sidestepping adversarial tuning and the risk of mode collapse. Diffusion training is easier to stabilise, yet it still preserves the fine details and textures essential for high-fidelity reconstruction. The same formulation also scales naturally to multi-resolution generation, delivering consistent quality across diverse visual inputs.

\subsection{Multi-Scale Diffusion Decoder}
To enable high-quality image synthesis with fast inference, we adopt a multi-scale conditional decoder based on Flow Matching~\cite{esser2024scaling}. Unlike conventional diffusion models that operate over hundreds of steps at a fixed resolution, our decoder progressively generates the image across multiple resolution scales. This hierarchical generation allows coarse structures to be first established at low resolution, and then refined with increasing spatial granularity, largely reducing the number of steps required at each stage.

\paragraph{Model Architecture.}
We adopt a transformer-based MMDiT (Multimodal Diffusion Transformer)~\cite{esser2024scaling} encoder–decoder architecture. MMDiT integrates diffusion modeling with a transformer backbone, enabling joint denoising and cross-modal conditioning across visual and semantic tokens. The encoder compresses an input image \( \mathbf{x}_0 \in \mathbb{R}^{3 \times H \times W} \) into a fixed-length sequence of 128 latent tokens, replacing the conventional spatial grid. After patch embedding with a patch size of 8, it produces 128 learned tokens of width 32, forming a compact context \( \mathbf{z} \in \mathbb{R}^{128 \times 32} \). 
This design compresses the original 1024 patch tokens into 128, decouples context length from image resolution, and keeps attention cost fixed while preserving global structure.

At each stage \( s \in \{1, \dots, S\} \), the decoder operates on a progressively upsampled spatial grid. Its inputs consist of the shared latent context \( z \) and a noisy feature map \( \mathbf{x}_t^{(s)} \), both represented as token grids. The decoder refines \( \mathbf{x}_t^{(s)} \) using cross-attention transformer layers conditioned on \( z \), while lightweight upsampling layers between stages progressively increase spatial resolution.

\paragraph{Multi-Stage Decoding Process.}
Our decoder takes a latent code $z$ and performs denoising in $S$ stages. Each stage $s \in \{1, \dots, S\}$ corresponds to a specific resolution $H_s \times W_s$ and iteratively denoises an image $x_t^{(s)}$ from an initial Gaussian sample to a refined final prediction $x_T^{(s)}$.







For the first stage, we initialize $x_0^{(1)} \sim \mathcal{N}(0, I)$. For later stages $s > 1$, we upsample the previous output $x_T^{(s-1)}$ to the new resolution and inject noise via
\begin{equation}
x_0^{(s)} \gets \alpha \cdot \text{Up}(x_T^{(s-1)}, H_s, W_s) + \beta \cdot \epsilon, \quad \epsilon \sim \mathcal{N}(0, I),
\end{equation}
where $\alpha$ and $\beta$ control the signal and noise respectively.

Within each stage, a small set of linearly spaced timesteps $\{t_i\}_{i=1}^{N_s}$ is used to numerically integrate a learned velocity field $v_t$ predicted by the decoder $\mu_\theta$. The overall process is guided by classifier-free guidance (CFG) with an adjustable guidance scale factor:
\begin{equation}
v_t \gets \mu_\theta(x, t, \emptyset) + \text{cfg\_scale} \cdot (\mu_\theta(x, t, z) - \mu_\theta(x, t, \emptyset)).
\end{equation}
Each timestep performs Euler integration:
\begin{equation}
x_{t_{i+1}} \gets x_{t_i} + (t_{i+1} - t_i) \cdot v_t.
\end{equation}
The output of the final stage $x_T^{(S)}$ is the reconstructed image.

\paragraph{Training Objective.} 
To train the multi-stage decoder, we sample an image $x$ and encode it into latent code $z = f_\phi(x)$. At each stage $s$, we define a forward process connecting the start and end states by downsampling $x$ at different scales:
\begin{align}
x_{t_0}^{(s)} &= t_s^0 \cdot \textit{Up}(\textit{Down}(x, 2^{s+1})) + (1 - t_s^0) \cdot \epsilon, \label{eq:train-start} \\
x_{t_1}^{(s)} &= t_s^1 \cdot \textit{Down}(x, 2^s) + (1 - t_s^1) \cdot \epsilon. \label{eq:train-end}
\end{align}
We then linearly interpolate between $x_{t_0}^{(s)}$ and $x_{t_1}^{(s)}$ to generate intermediate training pairs $x_t$, and train the model to predict the velocity field:
\begin{equation}
v_t = \frac{d x_t}{dt} = x_{t_1}^{(s)} - x_{t_0}^{(s)},
\end{equation}
minimizing the mean squared error:
\begin{equation}
\mathbb{E}_{s,t,x} \left[ \| \mu_\theta(x_t, \tau, z) - v_t \|_2^2 \right].
\end{equation}
The total training loss sums over all stages:
\begin{equation}
\mathcal{L}_{\text{total}} = \sum_{s=1}^S \mathcal{L}^{(s)}.
\end{equation}
Compared to traditional diffusion training that requires thousands of noise steps, our hierarchical training with a few steps per stage yields efficient and scalable learning.

\begin{table*}[t]
\centering
\footnotesize 
\renewcommand{\arraystretch}{1.12}

\begin{tabular}{l c c c c c}
\toprule

\textbf{Model} & \textbf{Num Tokens} & \textbf{rFID↓} & \textbf{PSNR↑} & \textbf{SSIM↑} & \textbf{Throughput (img/s)↑} \\
\midrule
\multicolumn{6}{c}{\textbf{Vector-Quantized Tokenizers}} \\[-2pt]
\midrule
TiTok-S-128 \cite{yu2024image}          & 128  & 1.71 & 17.52 & 0.437 & 7.31 \\
LlamaGen-16 \cite{sun2024autoregressive}         & 256  & 2.19 & 20.67 & 0.589 & 4.55 \\
Cosmos DI-16×16 \cite{agarwal2025cosmos}       & 256  & 4.40 & 19.98 & 0.536 & 9.55 \\
OpenMagViT-V2 \cite{luo2024open}      & 256  & 1.17 & 21.63 & 0.640 & 42.91 \\
ViT-VQGAN \cite{yu2021vector}           & 1024 & 1.28 & --    & --    & 94.96 \\
Cosmos DI-8×8 \cite{agarwal2025cosmos}         & 1024 & 0.87 & 24.82 & 0.763 & 26.28 \\
\midrule

\multicolumn{6}{c}{\textbf{Continuous Latent Tokenizers}} \\[-2pt]
\midrule
DC-AE \cite{chen2024deep}                & 64   & 0.77 & 23.93 & 0.766 & 60.66 \\
Cosmos CI-16×16 \cite{agarwal2025cosmos}      & 256  & 0.72 & 29.10 & 0.675 & 39.48 \\
TexTok-256 \cite{zha2025language}           & 256  & 0.73 & 24.45 & 0.668 & 18.18 \\
SD-VAE \cite{dai2018syntax}               & 1024 & 0.62 & 26.04 & 0.834 & 63.65 \\
Cosmos CI-8×8 \cite{agarwal2025cosmos}          & 1024 & 0.79 & 30.18 & 0.804 & 27.66 \\
\midrule

\multicolumn{6}{c}{\textbf{Diffusion-based Tokenizers}} \\[-2pt]
\midrule
FlowMo \cite{sargent2025flow}                & 256  & 0.95 & 22.07 & 0.649 & 1.44 \\
DiTo \cite{chen2025diffusion}                  & 256  & \textbf{0.78} & 24.10 & 0.706 & 0.19 \\
\rowcolor{gray!20}
Ours (After 1st stage) & 128 & 0.91 & 23.27 & 0.752 & 2.76 \\
\rowcolor{gray!20}
Ours (After 2nd stage) & 128 & 1.09 & \textbf{24.74} & \textbf{0.800} & \textbf{87.16} \\
\bottomrule

\end{tabular}
\caption{Tokenization comparison on ImageNet-1K at $256\times256$ resolution.  
Vector-quantized, continuous-latent, and diffusion tokenizers are evaluated by rFID, PSNR, SSIM, and throughput (images s$^{-1}$).  
Our multi-scale one-step decoder (the last line) approaches a comparable fidelity while running an order of magnitude faster than earlier diffusion tokenizers.}
\label{table:main-results}
\end{table*}

\subsection{Multi-Scale Distillation}

Despite the efficiency gains from multi-scale sampling, each stage of a diffusion decoder still requires multiple iterative denoising steps, which limits practical deployment. This is especially problematic at the final scale (e.g., 256 $\times$ 256 or higher), where high-resolution denoising dominates the runtime, often accounting for over half of the total decoding time. To address this bottleneck, we introduce an effective multi-scale distillation strategy that replaces costly iterative refinement with efficient single-step denoising at each scale, thereby enabling fast reconstruction with only minor sacrifice in perceptual quality.

\paragraph{Training Procedure.}
Our training pipeline consists of three components: a distilled-student model \( \theta \), a teacher model \( \psi \) with frozen weights, and a discriminator \( \phi \). The core idea is to distill the multi-step denoising process of the teacher into a \emph{coarse-to-fine student with the same architecture} that performs a single denoising step per spatial scale, drastically reducing inference cost while preserving fidelity.

Both the student and teacher decoders operate on the same latent representation $z$, which is obtained by encoding the clean ground-truth image \( \textit{x}_0 \) using a shared encoder:
\[
z = \textit{Encoder}(\textit{x}_0)
\]
During distillation training, the encoder is \textit{frozen}, and only the decoder parameters are updated. This focuses learning on transferring the denoising behavior of the teacher decoder to the student decoder.

The student decoder starts from a pure noise image:
\[
x_1 \sim \mathcal{N}(0, I)
\]
and performs \emph{one denoising step at each scale}, progressively refining the image from the coarsest resolution to the finest. Let \( S \) denote the number of spatial scales. At each stage \( s \in \{1, \dots, S\} \), the student produces an intermediate output \( \hat{x}^{(s)}_\theta \), with the final image denoted as \( \hat{x}^{(S)}_\theta \).
To supervise this process, we use the frozen teacher model \( \psi \) trained with full multi-step diffusion. We perturb the student’s final output by adding noise corresponding to a randomly sampled diffusion timestep \( t \in T_{\text{teacher}} \), producing:
\[
x_t = \alpha_t \hat{x}^{(S)}_\theta + \sigma_t \epsilon, \quad \epsilon \sim \mathcal{N}(0, I)
\]
We identify the corresponding spatial scale \( s_t \) associated with timestep \( t \), and let the teacher decode from scale \( s_t \) to \( S \), performing one denoising step per scale. The teacher’s outputs are denoted \( \hat{x}^{(s)}_\psi \) for \( s \geq s_t \).

This hierarchical supervision aligns the student with the teacher in both the final output and intermediate resolutions. Unlike conventional distillation that only matches terminal states, our design captures the teacher’s refinement process in a \textit{compact, one-step-per-scale} form.
We further enhance realism and perceptual quality through a discriminator and perceptual loss, detailed in the next section.

\paragraph{Loss Functions.}
Our training objective comprises three components: a multi-scale reconstruction loss to align the denoising trajectories, a perceptual loss to improve visual fidelity, and an adversarial loss to enhance realism. These jointly supervise the student decoder.

\vspace{2pt}
\textbf{Multi-Scale Reconstruction Loss.}
To ensure consistency with the teacher’s progressive denoising behavior, we compute a simple mean squared error between the student and teacher outputs across all resolutions, thereby aligning their predictions at each stage. Denoting the student and teacher outputs at scale \( s \) as \( \hat{x}^{(s)}_\theta \) and \( \hat{x}^{(s)}_\psi \), the loss becomes:
\[
\mathcal{L}_{\text{rec}} = \sum_{s = s_t}^{S} \left\| \hat{x}^{(s)}_\theta - \hat{x}^{(s)}_\psi \right\|_2^2
\]
This formulation encourages the student to mimic the teacher’s multi-stage refinement process.

\vspace{2pt}
\textbf{Perceptual Loss.}
Pixel-wise losses often fail to capture semantic similarity. To address this, we add a perceptual LPIPS loss that compares deep features extracted by a pre-trained VGG16 network. Given the student’s final prediction \( \hat{x}^{(S)}_\theta \) and the clean ground truth \( x_0 \), we define:
\[
\mathcal{L}_{\text{perc}} = \textit{LPIPS}(\hat{x}^{(S)}_\theta, x_0)
\]
This loss guides the student toward perceptually plausible outputs even in the presence of structural variations.

\vspace{2pt}
\textbf{Adversarial Loss.}
To further encourage realism, we incorporate an adversarial signal from a DINO-based patch-level discriminator \( D_\phi \). This discriminator extracts hierarchical ViT features and processes them using spectral and residual convolutions to classify real versus generated images. Compared to CNN-based discriminators, DINO features offer stronger perceptual gradients and improved convergence~\cite{caron2021emerging}. The generator and discriminator losses are:
\[
\mathcal{L}_{\text{adv}} = - \log D_\phi \left( \hat{x}^{(S)}_\theta \right)\]
\[\quad
\mathcal{L}_{\text{disc}} = - \log D_\phi \left( x_0 \right) - \log \left( 1 - D_\phi \left( \hat{x}^{(S)}_\theta \right) \right)
\]
\textbf{Final Objective.}
The full loss of student decoder training is given by:
\[
\mathcal{L}_{\text{total}} = \lambda_{\text{rec}} \mathcal{L}_{\text{rec}} + \lambda_{\text{perc}} \mathcal{L}_{\text{perc}} + \lambda_{\text{adv}} \mathcal{L}_{\text{adv}}
\]
where \( \lambda_{\text{rec}}, \lambda_{\text{perc}}, \lambda_{\text{adv}} \) are weighting coefficients that balance the contributions of each loss term.

\section{Experiments}
\subsection{Experiment Setup}

\begin{figure*}[t]
\centering
\includegraphics[width=1.6\columnwidth]
{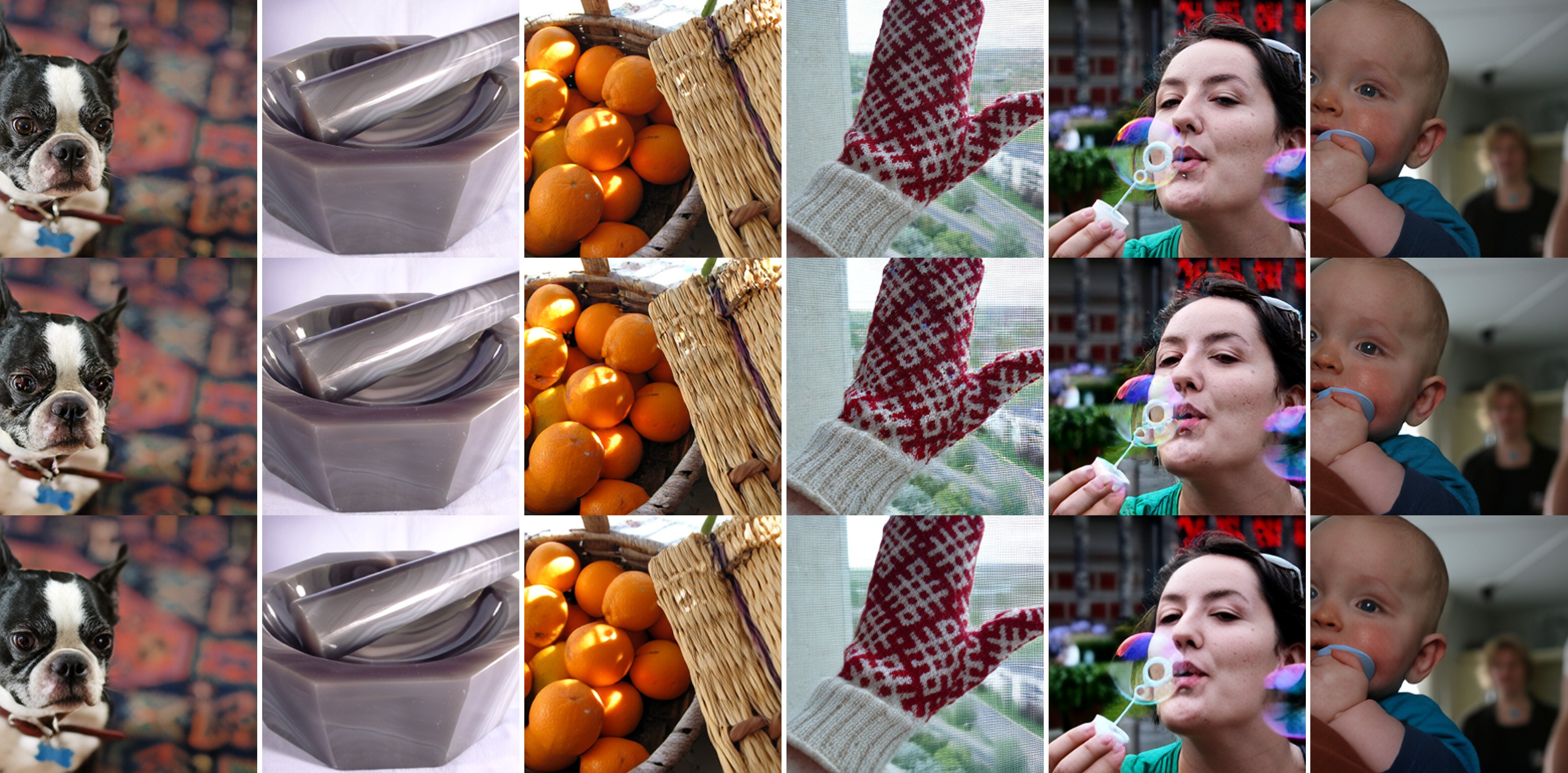} 
\caption{Representative reconstructions.  
Top: ground truth; middle: Stage-1 multi-scale model (30 steps/scale); bottom: Stage-2 distilled model (1 step/scale, 4 scales).  
The distilled decoder preserves visual fidelity while cutting the total denoising steps by \(\sim30\times\).}

\label{fig:recon_vis}
\end{figure*}

\paragraph{Dataset.} 
We conduct all experiments on the ImageNet-1K~\cite{deng2009imagenet} dataset. Following standard practice, we use the training split for model training and the validation split for evaluation. All input images are first center-cropped and then resized to a resolution of $256 \times 256$ along the shorter side. For evaluation, we report reconstruction FID (rFID)~\cite{heusel2017gans} computed between ground-truth images and their corresponding reconstructions. Additionally, we report PSNR~\cite{huynh2008scope}, SSIM~\cite{wang2004image}, and throughput (measured in images per second) to assess both quality and efficiency.

\paragraph{Tokenizer Architecture.} 
We train our proposed multi-scale diffusion tokenizer with a ViT-based encoder-decoder architecture. The encoder and decoder follow the MMDiT-D12~\cite{esser2024scaling} backbone with patch size 8, initialized from scratch and fully fine-tuned. The encoder transforms input images into 128 latent tokens of dimension 32, with cross-ROPE positional encoding applied. The decoder adopts a flow-matching-based conditional generation mechanism, conditioned on the latent representation.

\paragraph{Training Configuration.}
We adopt a two-stage training strategy on 8\,NVIDIA A100 GPUs.
\emph{Stage 1} jointly optimizes both the encoder and decoder for 200 epochs using the AdamW optimizer with a learning rate of $1\times10^{-4}$, $\beta_1$ of 0.9, $\beta_2$ of 0.95, no weight decay, and a cosine schedule that includes a 5-epoch warm-up. The batch size is set to 1024, the gradient clipping threshold is fixed at 1.0, and the model is trained using bfloat16 precision.
\emph{Stage 2} freezes the encoder and distills the decoder from its Stage 1 teacher in a single epoch, using the same optimizer settings. A lightweight discriminator is added in this stage and trained with a learning rate of $5\times10^{-5}$.

\subsection{Performance Analysis}
Table~\ref{table:main-results} evaluates our method on the \emph{ImageNet-1K validation set} at a resolution of $256\times256$ pixels, comparing it with two broad tokenizer families:  
(i) \emph{vector-quantised} and \emph{continuous latent} tokenizers that reconstruct an image with a \emph{single deterministic} decoder pass, and  
(ii) \emph{diffusion-based} tokenizers that require tens of iterative denoising steps.  
The latter group is renowned for perceptual quality yet suffers from markedly longer inference times because every sample traverses the decoder network many times, and the final $256\times256$ scale alone dominates overall latency.

\vspace{2pt}
\textbf{Efficiency limitations of diffusion tokenizers.} Diffusion tokenizers are inherently slow because each forward pass predicts noise for a single timestep, causing the total cost of image generation to grow linearly with the number of steps, typically around 25–50 at the highest resolution. In contrast, traditional tokenizers map the latents to the image space only once, making their computational cost essentially equivalent to a single network evaluation.

\vspace{2pt}
\textbf{Comparison with state-of-the-art methods.} Both \textit{DiTo}\cite{chen2025diffusion} and \textit{FlowMo}\cite{sargent2025flow}   represent the state-of-the-art among diffusion tokenizers, as reported in Table~\ref{table:main-results}. With 256 tokens, they achieve excellent rFID scores of 0.78 and 0.95, respectively. However, their throughputs are merely 0.19~$\text{img}\,\text{s}^{-1}$ for DiTo and 1.44~$\text{img}\,\text{s}^{-1}$ for FlowMo, highlighting the practical barrier imposed by iterative decoding.

Under the same 128-token latent budget, our model attains $87.16~\text{img}\,\text{s}^{-1}$,
corresponding to $459\times$ and $60\times$ speedups over DiTo and FlowMo, respectively.
This acceleration closes the entire gap to, and even surpasses, non-diffusion continuous tokenizers such as DC-AE and SD-VAE, which reach 60.66~img\,s$^{-1}$ and 63.65~img\,s$^{-1}$ respectively. In effect, we retain the stochastic, high-fidelity nature of diffusion while matching real-time performance that was previously reserved for feed-forward decoders.

\vspace{2pt}
\textbf{Quality trade-off.} A modest quality trade-off accompanies the speed-up, but the impact remains small: our rFID of 1.09 is just 0.14 above FlowMo and 0.31 above DiTo, yet it still clearly outperforms most non-diffusion tokenizers, including some that rely on four to eight times more tokens. PSNR (24.74) and SSIM (0.800) likewise remain near the top of the table. Figure~\ref{fig:recon_vis} visualises representative reconstructions produced by our method, further confirming the quantitative advantages discussed above.

\vspace{3pt}
Our multi-scale distilled diffusion decoder breaks the persistent quality–latency trade-off: it reduces inference time by \textbf{two to three orders of magnitude} relative to prior diffusion tokenizers, achieves throughput comparable to fast continuous methods, and maintains highly competitive reconstruction fidelity.

\begin{table*}[t]
\centering
\footnotesize 
\renewcommand{\arraystretch}{1.}

\begin{tabular}{>{\centering\arraybackslash}m{3.2cm} l c c c c}
\toprule
\textbf{Type} & \textbf{Method} & \textbf{rFID} $\downarrow$ & \textbf{PSNR} $\uparrow$ & \textbf{SSIM} $\uparrow$ & \textbf{Throughput (img/s)} $\uparrow$ \\
\midrule

\multirow{3}{*}{\textbf{Teacher Models}}
  & Single-scale (1 step)   & 2.22 & \textbf{25.96} & \textbf{0.83}  & 0.28 \\
  & Multi-scale (3 steps)   & 3.24 & 22.41 & 0.70  & 2.23 \\
  &\textbf{Multi-scale (4 steps)}   & \textbf{0.91} & 23.27 & 0.752 & \textbf{2.76} \\
\midrule

\multirow{3}{*}{\textbf{Distilled Students}}
  & Single-scale (1 step)   & 1.51 & 18.64 & 0.73  & \textbf{147.22} \\
  & Multi-scale (3 steps)   & 2.04 & 22.43 & 0.71  & 118.50 \\
  & \textbf{Multi-scale (4 steps)}   & \textbf{1.09} & \textbf{24.74} & \textbf{0.80}  & 87.16 \\
\bottomrule
\end{tabular}

\caption{
Comparison of throughput and reconstruction quality for teacher diffusion decoders and their distilled one-step students.
}
\label{table:scale-ablation}
\end{table*}


\subsection{Ablation Study}


\textbf{Effect of Multi-scale denoising.}  
To quantify the benefit of hierarchical sampling, we compare undistilled diffusion decoders that each use 120 denoising steps. With this identical budget, the single-scale baseline runs at only $0.28$ img/s (rFID $2.22$), whereas the four-stage counterpart reaches $2.76$ img/s and improves fidelity to rFID $0.91$, yielding an approximately $10\times$ speed-up by distributing work across coarse-to-fine scales. The four-stage design recovers finer detail by reserving a few steps for the final full-resolution pass, showing that deeper hierarchies boost both efficiency and quality without raising the step budget.

\vspace{2pt}
\textbf{Effect of Per-scale distillation.}
After distilling each stage to a single denoising step, the four-stage model attains $87.16$ img/s, which is more than $30\times$ faster than its teacher, while the rFID increases only from $0.91$ to $1.09$.  
Compared with a three-stage student, which is faster but noticeably less accurate, the four-stage distilled model offers a more favourable quality-latency operating point suitable for real-time applications.  
Detailed results are listed in Table~\ref{table:scale-ablation}.

\begin{figure}[t]
\centering
\includegraphics[page=1, clip, trim=20 0 310 0, width=0.85\linewidth]{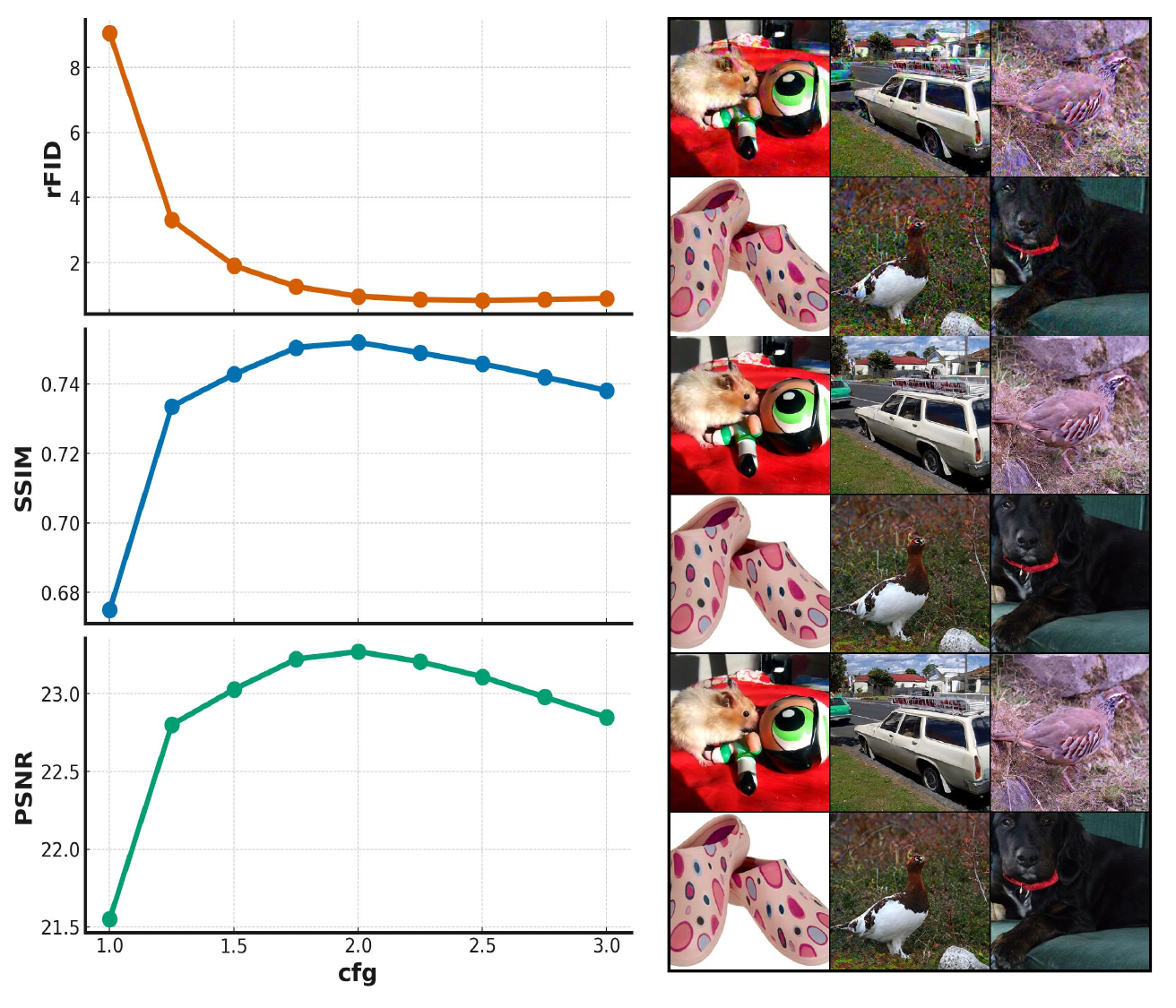} 
\caption{Effect of cfg on Stage-1 training after 200 epochs.
Left: rFID, SSIM, and PSNR as cfg varies.
Right: Reconstruction examples for cfg = 1, 2, 3 (top to bottom). A cfg value around 2 offers the best balance of fidelity and perceptual quality.}

\label{fig:recon_vis4}
\end{figure}

\vspace{2pt}
\textbf{Hyper-parameter Sensitivity.}
We first examine the effect of cfg\cite{ho2022classifier} during Stage-1 training (Figure \ref{fig:recon_vis4}). Increasing cfg from 1.0 to 2.0 consistently improves reconstruction fidelity: rFID drops sharply while SSIM and PSNR reach their maximum at cfg = 2. Higher values lead to mild degradation and occasional artifacts. Based on these trends, we adopt cfg of 2 and use the corresponding trained checkpoint as the teacher for Stage-2 distillation.

We then vary the perceptual loss weight $\lambda_{\text{perc}}\in\{0.1, 0.5,1.0,2.0\}$ and the guidance scale $\mathrm{cfg}\in\{1,2\}$ on the distillation stage.  
As shown in Table~\ref{tab:stage2_ablation}, a weight above $1.0$ consistently reduces fidelity, indicating that an overly strong perceptual term erodes fine details.  
Increasing $\mathrm{cfg}$ from $1$ to $2$ likewise degrades rFID, PSNR, and SSIM because the single-step student cannot fully remove the added high-frequency noise.
With $\mathrm{cfg}$ set to 1, the four-scale student is best at a $\lambda_{\text{perc}}$ of 0.5, reaching an rFID of 1.09, PSNR of 24.74, and SSIM of 0.80.  
The three-scale variant instead prefers $\lambda_{\text{perc}}$ of 1.0 and reaches the highest throughput at $130,\text{img},\text{s}^{-1}$ with lower fidelity (rFID 2.04).
Overall, these trends indicate that a four-stage hierarchy with moderate perceptual weighting and $\mathrm{cfg}$ of 1 provides the best quality–latency balance. We therefore adopt a one-step, four-scale decoder with $\lambda_{\text{perc}}$ of 0.5 and $\mathrm{cfg}$ of 1 as the default setting, achieving real-time decoding while remaining within 0.2 rFID of the teacher.

\begin{table}[t]
\centering
\footnotesize 
\renewcommand{\arraystretch}{1.0}

\begin{tabular}{>{\centering\arraybackslash}m{2.8cm} c c c c}
\toprule
\textbf{number of scale / cfg} &
$\lambda_{\text{perc}}$ &
\textbf{rFID} $\downarrow$ &
\textbf{PSNR} $\uparrow$ &
\textbf{SSIM} $\uparrow$ \\
\midrule

\multirow{4}{*}{\textbf{3 / 1.0}}
 & 0.1 & 2.34 & 21.63 & 0.69 \\
 & 0.5 & 2.19 & 21.97 & 0.69 \\
 & 1.0 & 2.04 & 22.43 & 0.71 \\
 & 2.0 & 2.22 & 22.62 & 0.71 \\
\midrule

\multirow{4}{*}{\textbf{3 / 2.0}}
 & 0.1 & 3.44 & 21.25 & 0.64 \\
 & 0.5 & 3.40 & 21.59 & 0.65 \\
 & 1.0 & 3.06 & 21.80 & 0.65 \\
 & 2.0 & 3.55 & 21.96 & 0.66 \\
\midrule

\multirow{4}{*}{\textbf{4 / 1.0}}
 & 0.1 & 1.17 & 24.26 & 0.78 \\ 
 & 0.5 & \textbf{1.09} & 2\textbf{4.74} & \textbf{0.80} \\
 & 1.0 & 1.25 & 23.11 & 0.76 \\
 & 2.0 & 1.40 & 23.43 & 0.76 \\
\midrule

\multirow{4}{*}{\textbf{4 / 2.0}}
 & 0.1 & 1.72 & 22.34 & 0.72 \\
 & 0.5 & 1.40 & 22.03 & 0.73 \\
 & 1.0 & 1.51 & 22.40 & 0.74 \\
 & 2.0 & 1.68 & 22.65 & 0.74 \\
\bottomrule
\end{tabular}

\caption{
Impact of perceptual-loss weight $\lambda_{\text{perc}}$ and classifier-free guidance (cfg)
on reconstruction quality across student decoders.
}
\label{tab:stage2_ablation}
\end{table}

\section{Conclusion}\label{sec:conclusion}

We present a novel two-stage accelerator framework for diffusion-based image tokenizers.  
A multi-scale sampler reduces the spatial complexity, and one-step per-scale distillation collapses dozens of denoising iterations into a single forward pass.
On ImageNet, our model achieves \(87\,\text{img}\,\text{s}^{-1}\) with only a \(0.18\) rFID increase, achieving a 2 to 3-order-of-magnitude speedup over standard diffusion decoders while maintaining competitive reconstruction fidelity.

{
    \small
    \bibliographystyle{ieeenat_fullname}
    \bibliography{main}
}


\end{document}